\documentclass[conference]{IEEEtran}
\IEEEoverridecommandlockouts
\usepackage{cite}
\usepackage{amsmath,amssymb,amsfonts}
\usepackage{algorithmic}
\usepackage{graphicx}
\usepackage{textcomp}
\usepackage{xcolor}
\usepackage{booktabs}
\usepackage{array}
\usepackage{tabularx}
\usepackage{adjustbox}
\usepackage{multirow}
\usepackage{colortbl}
\usepackage{subcaption}
\usepackage{caption}
\usepackage{textcase}
\definecolor{lightgray}{gray}{0.9}

\renewcommand{\thetable}{\Roman{table}}
\DeclareCaptionTextFormat{uppercase}{\MakeTextUppercase{#1}}
\def\BibTeX{{\rm B\kern-.05em{\sc i\kern-.025em b}\kern-.08em
    T\kern-.1667em\lower.7ex\hbox{E}\kern-.125emX}}
\begin{document}

\title{ArmFormer: Lightweight Transformer Architecture for Real-Time Multi-Class Weapon Segmentation and Classification}

\author{\IEEEauthorblockN{Akhila Kambhatla\textsuperscript{*,\dag,1,2}, Taminul Islam\textsuperscript{\dag,1,2}, and Khaled R Ahmed\textsuperscript{1,2}}
\IEEEauthorblockA{\textsuperscript{1}School of Computing, Southern Illinois University, Carbondale, IL, USA\\
\textsuperscript{2}BASE Lab, Southern Illinois University, Carbondale, IL, USA\\
\{akhila.kambhatla, taminul.islam, khaled.ahmed\}@siu.edu}
\thanks{\textsuperscript{*}Corresponding author.}
%\thanks{\textsuperscript{\dag}Akhila Kambhatla and Taminul Islam contributed equally to this work.}
}

\maketitle

\begin{abstract}
The escalating threat of weapon-related violence necessitates automated detection systems capable of pixel-level precision for accurate threat assessment in real-time security applications. Traditional weapon detection approaches rely on object detection frameworks that provide only coarse bounding box localizations, lacking the fine-grained segmentation required for comprehensive threat analysis. Furthermore, existing semantic segmentation models either sacrifice accuracy for computational efficiency or require excessive computational resources incompatible with edge deployment scenarios. This paper presents ArmFormer, a lightweight transformer-based semantic segmentation framework that strategically integrates Convolutional Block Attention Module (CBAM) with MixVisionTransformer architecture to achieve superior accuracy while maintaining computational efficiency suitable for resource-constrained edge devices. Our approach combines CBAM-enhanced encoder backbone with attention-integrated hamburger decoder to enable multi-class weapon segmentation across five categories: handgun, rifle, knife, revolver, and human. Comprehensive experiments demonstrate that ArmFormer achieves state-of-the-art performance with 80.64\% mIoU and 89.13\% mFscore while maintaining real-time inference at 82.26 FPS. With only 4.886G FLOPs and 3.66M parameters, ArmFormer outperforms heavyweight models requiring up to 48× more computation, establishing it as the optimal solution for deployment on portable security cameras, surveillance drones, and embedded AI accelerators in distributed security infrastructure.
\end{abstract}


\section{Introduction}

The escalating threat of firearm violence poses a critical challenge to public safety, with over 42,155 firearm-related deaths reported in the United States in 2024-2025 \cite{cdc}. Traditional security infrastructure relying on human surveillance and manual threat assessment has proven inadequate for addressing modern security challenges, particularly in high-traffic environments such as airports, educational institutions, and urban centers where potential threats exceed human operational capabilities. Surveillance systems require continuous human monitoring, yet concentration deteriorates to 83\%, 84\%, and 64\% after one hour when monitoring 4, 9, and 16 screens respectively \cite{sur_cam}, underscoring the need for automated threat detection systems.

Current weapon detection methods predominantly employ object detection frameworks providing bounding box localizations \cite{dong_yolov8, datacamp}, yet fail to deliver pixel-level precision necessary for accurate threat assessment in complex scenarios involving occlusions \cite{tiwari}, varying illumination \cite{Rao}, and multiple weapon types \cite{nadeem, survey-1}. While YOLO-based approaches demonstrate effectiveness in general object detection\cite{ak2}, they struggle with small and partially occluded objects \cite{dong_yolov8, ingle}, which are common in weapon detection scenarios. Semantic segmentation addresses these limitations by providing pixel-level classification and precise boundary delineation essential for accurate threat assessment \cite{ibm, keymakr}.

Recent transformer-based semantic segmentation architectures such as SegFormer \cite{segformer} have achieved superior performance in general computer vision tasks, demonstrating state-of-the-art results on benchmarks like ADE20K and Cityscapes with 5× speed improvements \cite{hugface}. However, these advances remain largely unexplored for security-critical applications requiring specialized multi-class weapon identification, with current research predominantly relying on traditional CNN-based approaches \cite{egia20,kaya,santos}. The absence of transformer-based methodologies in weapon detection is particularly notable given their demonstrated effectiveness in modeling complex spatial relationships and extracting multi-resolution features \cite{bai}, suggesting significant untapped potential for enhancing both accuracy and operational efficiency.

Despite revolutionizing semantic segmentation tasks, existing transformer architectures exhibit fundamental deficiencies when applied to weapon detection: excessive computational overhead incompatible with real-time requirements \cite{mmseg,swin, vit}, lack of specialized attention mechanisms for weapon classification, and performance degradation when distinguishing morphologically similar weapon categories under challenging environmental conditions \cite{Rao,tiwari,nadeem}. Multi-class weapon classification is essential for threat assessment, as distinguishing weapon types (handgun, rifle, knife, revolver) enables security systems to evaluate threat levels, operational ranges, and deploy proportionate countermeasures—whereas binary classification offers insufficient information for nuanced decision-making in critical security situations. In response to these limitations, this research introduces ArmFormer. This transformer-based segmentation architecture employs strategic Convolutional Block Attention Module (CBAM) \cite{cbam} integration across encoder-decoder pathways to enhance multi-class weapon recognition while preserving computational tractability. Our methodology leverages Google Open Images and IMFDB database \cite{imfdb} to enable granular pixel-level classification of weapon classes and human subjects. Experimental validation demonstrates that ArmFormer achieves superior performance compared to established baseline models while maintaining the inference efficiency required for real-world security deployment.

\section{Related Work}
Weapon detection methodologies have evolved from traditional sensor-based approaches to sophisticated deep learning architectures, yet achieving pixel-level precision with real-time efficiency for multi-class weapon segmentation remains an open challenge.

\subsection{Weapon Detection: From Traditional to Deep Learning Approaches}
Early weapon detection systems employed traditional techniques including radar \cite{microwave}, millimeter wave imaging \cite{3d}, and thermal/infrared sensors \cite{imagefusion} for security screening in airports and public venues, with TSA intercepting 6,678 firearms at checkpoints in 2024 \cite{tsa}. These methods utilized pattern matching \cite{auto_concealed}, density descriptors \cite{comp3d}, and cascade classifiers \cite{milli}, but suffered from high false positive rates due to object orientation variability and required continuous human monitoring of surveillance feeds \cite{ak1}.

The advent of deep learning transformed weapon detection through automated feature extraction. Classical computer vision approaches using hand-crafted descriptors such as HOG \cite{hog} and SIFT \cite{sift} with conventional classifiers were succeeded by CNN-based architectures including VGGNet \cite{vggnet} and ResNet \cite{resnet} for weapon recognition \cite{cnn}. Single-stage object detection frameworks, particularly YOLO variants (YOLOv3, YOLOv4, YOLOv8), have demonstrated superior real-time performance \cite{yolo, yolov3, yolov4}. However, these methods present inherent limitations: coarse bounding box localization lacking pixel-level granularity, diminished accuracy on small-scale weapons, and insufficient robustness against partial occlusions \cite{survey2}. Furthermore, CNNs' limited receptive fields restrict their capacity to model long-range spatial dependencies essential for comprehensive scene understanding \cite{imagenet, simon}, while transformers address these limitations through self-attention mechanisms enabling global spatial relationship modeling \cite{carion}.

\subsection{Transformer-based Semantic Segmentation and Attention Mechanisms}
Vision Transformers (ViTs) \cite{vit} have revolutionized dense prediction tasks by establishing self-attention as an effective alternative to convolutions for pixel-level classification. Semantic segmentation has progressed from FCNs \cite{fully} and U-Net \cite{unet} to transformer-based frameworks like SegFormer\cite{segformer}, Segmenter\cite{segmenter}, Swin\cite{swin}, achieving superior performance through hierarchical feature representation with efficient mix-FFN designs and overlapped patch embeddings \cite{pyramid}. While transformer-based approaches have shown promise for weapon detection through DETR-based localization frameworks \cite{survey3}, their application to pixel-level weapon segmentation remains largely unexplored, with computational efficiency challenges limiting deployment on resource-constrained security infrastructure.

Attention mechanisms have evolved from Squeeze-and-Excitation (SE) blocks to sophisticated spatial-channel integration schemes enhancing feature discriminability \cite{hu18, li19}. The Convolutional Block Attention Module (CBAM) represents a significant advancement, systematically combining channel and spatial attention pathways for sequential feature refinement, demonstrating enhanced selectivity and superior multi-scale object handling \cite{residual,survey4,alom}. Recent weapon detection research has shifted toward targeted attention mechanisms rather than complete transformer architectures, as full self-attention computations prove too expensive for real-time security systems. However, attention mechanism exploration remains limited in security-critical applications, with most research focusing on general-purpose vision tasks rather than specialized weapon detection requirements.

\subsection{Research Gap and Motivation}
Current transformer-based models exhibit critical deficiencies for weapon detection: excessive computational overhead incompatible with real-time deployment, lack of domain-specific optimizations for distinguishing morphologically similar weapon categories, and suboptimal attention mechanisms designed for general vision rather than security-critical applications \cite{mmseg,swin,vit,Rao,tiwari,nadeem}. The fundamental challenge lies in balancing pixel-level precision with computational efficiency, as existing models sacrifice inference speed for accuracy. No prior work has combined transformers with pixel-level semantic segmentation specifically for multi-class weapon classification, despite transformers' proven effectiveness in general segmentation tasks \cite{segformer,segmenter} and weapon detection exploration through CNN-based \cite{kaya, egia20} or DETR-based \cite{detr} approaches. ArmFormer addresses this gap by delivering precise pixel-level weapon classification while maintaining computational tractability essential for operational security systems. 

\begin{table}[htbp]
\centering
\caption{Dataset Samples Distribution (Count)}
\label{tab:dataset}
\footnotesize
\begin{adjustbox}{width=\columnwidth,center}
\begin{tabular}{lcccc}
\toprule
\textbf{Class } & \textbf{Test} & \textbf{Train} & \textbf{Validation} & \textbf{Total}\\
\midrule
Handgun  &   215  &  1506    &   440  &  2161   \\
Human    &    133   &   928     &    269   &  1330 \\
Knife    &   145   &  1017   &    288   &  1450 \\
Revolver  &   172  &   1280   &   345  &   1797 \\
Rifle   &   167  &   1115   &  328   &  1610 \\
\bottomrule
\end{tabular}
\end{adjustbox}
\end{table}

\section{Dataset and Pre-Processing}
\begin{figure}[htbp]
\centering
\includegraphics[width=\columnwidth]{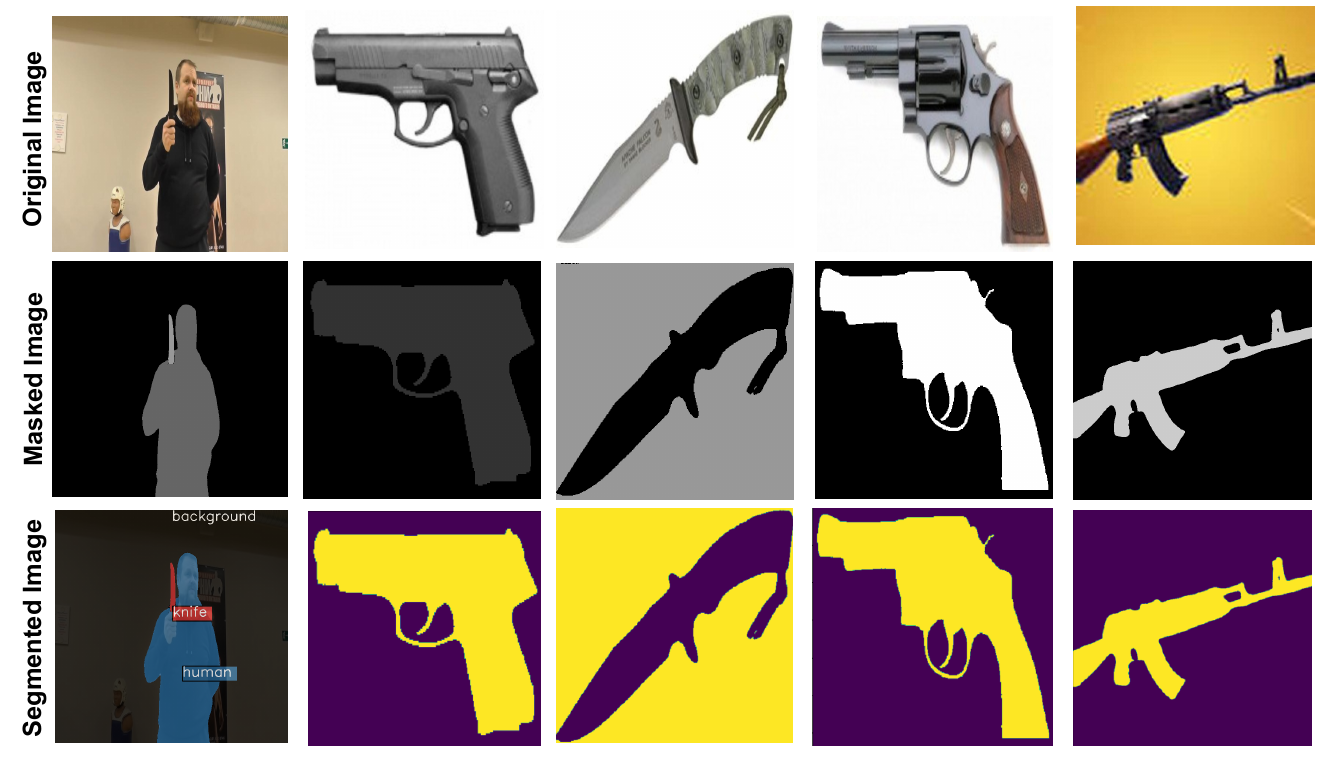}
\caption{Visualization of original images, masked annotations, and segmented results for each weapon category including handgun, human, knife, revolver, and rifle.}
\label{fig:dataset_samples}
\end{figure}

A lack of a consistent dataset for firearm segmentation and recognition prompted the development of a unique dataset consisting of 8097 images curated from Google Open Images and the IMFDB\cite{imfdb} database. High-quality weapon images with diverse perspectives were carefully selected to ensure effective real-world detection performance. Preprocessing involved systematic removal of background noise and irrelevant elements from each image using computational tools to enhance dataset quality and training relevance. The dataset annotation process employs a sophisticated semi-automated approach utilizing the Segment Anything Model 2 (SAM2) framework \cite{sam2} for multi-class object segmentation and has demonstrated robust multi-class segmentation performance across diverse domains \cite{islam2025weedswin}. The pipeline standardizes input image to $640 X 640 $ and implements an interactive annotation strategy that combines minimal human supervision with automated propagation across image sequences. SAM2's video/image prediction capabilities propagate the initial annotations across the entire image sequence, maintaining object identity and segmentation consistency throughout temporal sequences. The model generates both individual class-specific masks and combined multi-class masks. Each class is assigned unique identifiers and corresponding grayscale values for mask generation \textit{(Handgun: 51, Human: 102, Knife: 153, Rifle: 204, Revolver: 255)}, enabling precise class differentiation in the ground truth annotations while maintaining computational efficiency. Table [\ref{tab:dataset}] presents the distribution of samples across training, testing, and validation sets for each weapon class, while Figure \ref{fig:dataset_samples} illustrates representative examples from each class alongside their corresponding ground truth masks and segmented outputs.

\section{Methodology}
\subsection{ArmFormer Architecture}
\begin{figure*}[t]
\centering
\includegraphics[width=\textwidth]{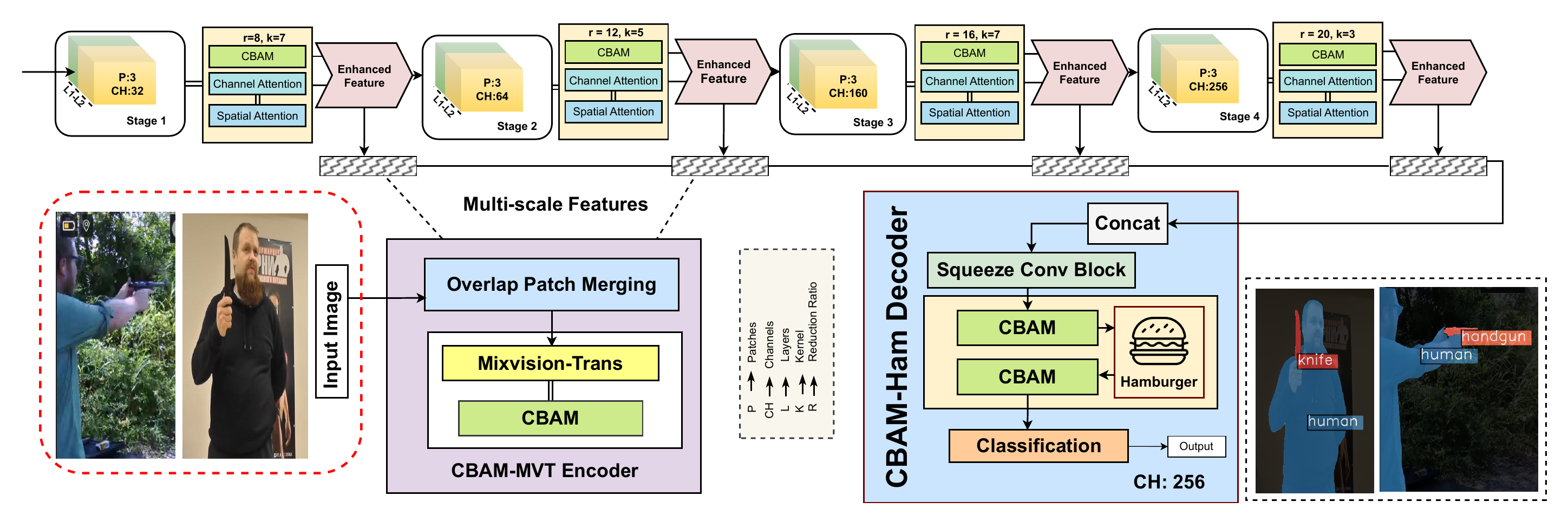}
\caption{ArmFormer Architecture Overview: The framework consists of a four-stage CBAM-enhanced MixVisionTransformer encoder backbone with progressive spatial resolution reduction and channel expansion, followed by a CBAM-integrated lightweight hamburger decoder head. The encoder processes input images through hierarchical stages (Stage 1-4) with progressive patch embeddings (P-3) and dual attention mechanisms (CBAM with Channel and Spatial Attention). Multi-scale features from all stages undergo overlap patch merging and are fused through the CBAM-MVT Encoder with Mixvision-Trans and CBAM modules. The decoder employs a squeeze convolution block, dual CBAM modules configured as a hamburger structure, and a classification head to generate pixel-level segmentation predictions for multi-class weapon detection.}
\label{fig:armformer_architecture}
\end{figure*}

We propose ArmFormer, a lightweight transformer-based semantic segmentation framework that strategically integrates Convolutional Block Attention Module (CBAM) \cite{cbam} with MixVisionTransformer \cite{yu2023mix} architecture to achieve superior feature representation capabilities for real-time multi-class weapon detection. As illustrated in Figure \ref{fig:armformer_architecture}, the architecture employs a hierarchical encoder-decoder paradigm with attention mechanisms integrated at both encoder and decoder stages to enhance feature discrimination while maintaining computational efficiency suitable for edge deployment.

\subsubsection{CBAM Attention Mechanism}
CBAM serves as the core attention mechanism in ArmFormer, enhancing feature representation through sequential channel and spatial attention refinement. Given an input feature map $\mathbf{F} \in \mathbb{R}^{C \times H \times W}$ with $C$ channels, height $H$, and width $W$, CBAM first applies channel attention to identify important feature channels, followed by spatial attention to localize informative regions.

The channel attention module exploits inter-channel relationships by aggregating spatial information through both average and max pooling operations. The channel attention map $\mathbf{M}_c \in \mathbb{R}^{C \times 1 \times 1}$ is computed as:
\begin{equation}
\mathbf{M}_c = \sigma(\text{MLP}(\text{AvgPool}(\mathbf{F})) + \text{MLP}(\text{MaxPool}(\mathbf{F})))
\end{equation}
where $\sigma$ denotes sigmoid activation and MLP is a shared multi-layer perceptron with reduction ratio $r=16$ that compresses and expands channel dimensions. The channel-refined feature is obtained as $\mathbf{F}' = \mathbf{M}_c \otimes \mathbf{F}$.

Following channel attention, the spatial attention module computes attention weights across spatial locations. The spatial attention map $\mathbf{M}_s \in \mathbb{R}^{1 \times H \times W}$ is generated by:
\begin{equation}
\mathbf{M}_s = \sigma(\text{Conv}^{7 \times 7}([\text{AvgPool}_c(\mathbf{F}'); \text{MaxPool}_c(\mathbf{F}')]))
\end{equation}
where pooling operates along the channel dimension, and the concatenated descriptors are processed through a $7 \times 7$ convolution with kernel size $k=7$. The final CBAM-refined feature output is:
\begin{equation}
\mathbf{F}_{out} = \mathbf{M}_s \otimes \mathbf{F}' = \mathbf{M}_s \otimes (\mathbf{M}_c \otimes \mathbf{F})
\end{equation}
This sequential attention mechanism allows the network to adaptively emphasize discriminative features while suppressing irrelevant background information, crucial for accurate weapon detection in complex security scenarios.

\subsubsection{CBAM-Enhanced Encoder Backbone}
The encoder backbone implements a CBAM-enhanced MixVisionTransformer architecture consisting of four hierarchical stages with progressive spatial resolution reduction and channel expansion to capture multi-scale contextual features. The input image $\mathbf{I} \in \mathbb{R}^{3 \times H_0 \times W_0}$ is processed through hierarchical stages with the following channel configurations: Stage 1 produces features with 32 channels, Stage 2 expands to 64 channels, Stage 3 increases to 160 channels, and Stage 4 reaches 256 channels, corresponding to spatial resolutions of $H_0/4$, $H_0/8$, $H_0/16$, and $H_0/32$ respectively.

Each stage employs overlapped patch embeddings with stride 4 and kernel size 7 to partition the feature map into overlapping patches, preserving local continuity while enabling efficient self-attention computation. The transformer blocks within each stage utilize efficient self-attention mechanisms that model long-range dependencies, combined with mix-FFN (Mix Feed-Forward Network) \cite{aftabi2025feed} layers that incorporate depth-wise convolutions for position encoding. Following each transformer stage, CBAM modules with uniform parameters (reduction ratio $r=16$, spatial kernel size $k=7$) are strategically inserted to enhance the extracted features through channel and spatial attention refinement.

The hierarchical design enables the encoder to capture both fine-grained local details from early stages and high-level semantic information from deeper stages, producing multi-scale feature representations $\{\mathbf{F}_1, \mathbf{F}_2, \mathbf{F}_3, \mathbf{F}_4\}$ that are essential for accurate dense prediction. The CBAM integration at each stage allows the network to focus on weapon-specific features while maintaining computational efficiency, with the attention mechanisms learning to emphasize relevant spatial regions and informative feature channels for multi-class weapon discrimination.

\subsubsection{CBAM-Integrated Decoder Head}
The decoder head employs a lightweight hamburger architecture with strategic dual CBAM integration to efficiently aggregate multi-scale features and generate pixel-level segmentation predictions. Multi-scale features $\{\mathbf{F}_1, \mathbf{F}_2, \mathbf{F}_3, \mathbf{F}_4\}$ from all encoder stages undergo overlap patch merging, where features are first aligned to unified spatial dimensions through bilinear interpolation, then concatenated along the channel dimension to form a comprehensive multi-scale representation.

The hamburger module performs efficient global context modeling through matrix decomposition, enabling the decoder to capture long-range dependencies with significantly reduced computational overhead compared to full self-attention mechanisms. This is achieved by decomposing the feature map into lower-rank matrices that preserve global contextual information while maintaining linear complexity with respect to spatial resolution.

CBAM attention is applied at two strategic positions within the decoder pathway. The pre-hamburger CBAM ($\text{CBAM}_1$) refines the concatenated multi-scale features by enhancing channel-wise relationships and emphasizing informative spatial regions before global context aggregation. Following the hamburger module, post-hamburger CBAM ($\text{CBAM}_2$) further enhances the globally-contextualized features by applying sequential attention refinement. The complete decoder processing pipeline is formulated as:
\begin{equation}
\begin{split}
\mathbf{S} = \text{Conv}_{cls}(&\text{CBAM}_2(\text{Ham}(\\
&\text{CBAM}_1(\text{Concat}(\mathbf{F}_1, \mathbf{F}_2, \mathbf{F}_3, \mathbf{F}_4)))))
\end{split}
\end{equation}
where $\text{Ham}$ denotes the hamburger module for global context aggregation, $\text{Conv}_{cls}$ is the classification convolution layer, and $\mathbf{S} \in \mathbb{R}^{N \times H \times W}$ represents the final segmentation prediction with $N$ classes including handgun, rifle, knife, revolver, and human.

The complete ArmFormer architecture integrates all components through end-to-end training using pixel-wise cross-entropy loss \cite{mao2023cross}:
\begin{equation}
\mathcal{L} = -\frac{1}{HW}\sum_{i=1}^{H}\sum_{j=1}^{W}\sum_{n=1}^{N} y_{ij}^{(n)} \log(\hat{y}_{ij}^{(n)})
\end{equation} 
where $y_{ij}^{(n)}$ represents the ground truth label and $\hat{y}_{ij}^{(n)}$ denotes the predicted probability for class $n$ at spatial location $(i,j)$. This end-to-end optimization allows joint learning of attention mechanisms and segmentation performance, enabling the network to adaptively focus on task-specific features tailored for multi-class weapon detection while maintaining the lightweight computational profile essential for real-time edge deployment.

\section{Experimental Results}
This section focuses on conducting comprehensive experiments to evaluate the performance of our proposed ArmFormer against state-of-the-art semantic segmentation models across multiple dimensions including segmentation accuracy, computational efficiency, and per-class performance. The evaluation encompasses both lightweight models designed for edge deployment (CGNet\cite{wu2020cgnet}, HrNet\cite{SunXLW19}) and heavyweight architectures optimized for accuracy (EncNet\cite{Zhang_2018_CVPR}, ICNet\cite{zhao2018icnet}, Uppernet\_swin\cite{xiao2018unified}), providing a thorough assessment of ArmFormer's competitive positioning across the efficiency-accuracy spectrum.

\subsection{Comparison with State-of-the-Art Methods}

Table \ref{tab:comparison_overall} presents the comprehensive performance comparison of our proposed ArmFormer against eight state-of-the-art semantic segmentation architectures. The evaluation metrics include mean Intersection over Union (mIoU), mean Accuracy (mAcc), mean F-score (mFscore), computational complexity (FLOPs in Giga operations), model parameters (in Millions), and inference speed (Frames Per Second).

\begin{table}[htbp]
\centering
\caption{Performance Comparison}
\label{tab:comparison_overall}
\footnotesize
\begin{adjustbox}{width=\columnwidth,center}
\begin{tabular}{lcccccc}
\toprule
\textbf{Model} & \textbf{FLOPs} & \textbf{Params} & \textbf{Speed} & \textbf{mIoU} & \textbf{mAcc} & \textbf{mFscore} \\
 & \textbf{(G)} & \textbf{(M)} & \textbf{(FPS)} & \textbf{(\%)} & \textbf{(\%)} & \textbf{(\%)} \\
\midrule
\rowcolor{lightgray}
\textbf{ArmFormer (Ours)} & \textbf{4.886} & \textbf{3.66} & \textbf{82.26} & \textbf{80.64} & \textbf{88.28} & \textbf{89.13} \\
CGNet \cite{wu2020cgnet} & 3.452 & 0.493 & 90.49 & 64.30 & 74.96 & 77.08 \\
ICNet\cite{zhao2018icnet} & 15.434 & 47.52 & 140.88 & 74.16 & 82.41 & 84.63 \\
Segmenter\cite{segmenter} & 12.266 & 6.685 & 74.15 & 31.46 & 41.46 & 50.86 \\
Uppernet\_swin\cite{xiao2018unified} & 236.0 & 58.94 & 38.97 & 70.90 & 79.53 & 82.56 \\
PspNet\cite{zhao2017pspnet} & 18.14 & 4.571 & 56.96 & 66.20 & 75.12 & 77.95 \\
EncNet\cite{Zhang_2018_CVPR} & 54.56 & 12.52 & 90.78 & 77.65 & 81.65 & 80.63 \\
HrNet\cite{SunXLW19} & 6.52 & 1.87 & 64.92 & 69.24 & 78.31 & 81.55 \\
\bottomrule
\end{tabular}
\end{adjustbox}
\end{table}

Table \ref{tab:comparison_classwise} provides detailed per-class IoU performance across all weapon categories and human detection, revealing the model's capability to handle diverse object types in security-critical scenarios.

\begin{table}[htbp]
\centering
\caption{Per-Class IoU Performance Comparison (\%)}
\label{tab:comparison_classwise}
\footnotesize
\begin{adjustbox}{width=\columnwidth,center}
\begin{tabular}{lccccc}
\toprule
\textbf{Model} & \textbf{Handgun} & \textbf{Human} & \textbf{Knife} & \textbf{Rifle} & \textbf{Revolver} \\
\midrule
\rowcolor{lightgray}
\textbf{ArmFormer (Ours)} & \textbf{82.24} & \textbf{67.22} & \textbf{80.81} & \textbf{83.87} & \textbf{80.43} \\
CGNet\cite{wu2020cgnet}& 71.13 & 32.92 & 66.42 & 72.67 & 61.90 \\
ICNet\cite{zhao2018icnet} & 78.72 & 50.12 & 76.83 & 77.46 & 75.65 \\
Segmenter\cite{strudel2021segmenter} & 33.19 & 7.41 & 59.24 & 11.56 & 22.02 \\
Uppernet\_swin\cite{xiao2018unified} & 74.88 & 52.10 & 73.56 & 74.23 & 66.52 \\
PspNet\cite{zhao2017pspnet} & 81.82 & 55.81 & 77.45 & 29.80 & 73.18 \\
EncNet\cite{Zhang_2018_CVPR} & 83.31 & 58.77 & 81.83 & 80.17 & 72.12 \\
HrNet\cite{SunXLW19} & 72.78 & 58.08 & 61.83 & 70.01 & 69.08 \\
\bottomrule
\end{tabular}
\end{adjustbox}
\end{table}

\subsection{Performance Analysis and Model Superiority}

ArmFormer achieves state-of-the-art performance with 80.64\% mIoU and 89.13\% mFscore, outperforming all baseline models including heavyweight architectures. Our model surpasses EncNet\cite{Zhang_2018_CVPR} (77.65\% mIoU) despite using 11.2× fewer FLOPs (4.886G vs 54.56G) and 3.4× fewer parameters (3.66M vs 12.52M), and significantly exceeds Uppernet\_swin\cite{xiao2018unified} (70.90\% mIoU) while being 48× more computationally efficient. With 82.26 FPS inference speed, ArmFormer maintains real-time processing capabilities essential for security applications. While ICNet\cite{zhao2018icnet} achieves higher FPS (140.88), it sacrifices 6.48\% mIoU, and CGNet\cite{wu2020cgnet} prioritizes speed (90.49 FPS) but underperforms by 16.34\% mIoU, making both unsuitable for accurate threat assessment.

Table \ref{tab:comparison_classwise} demonstrates ArmFormer's robust multi-class detection, achieving the highest performance in four out of five classes: Handgun (82.24\%), Rifle (83.87\%), Knife (80.81\%), and Revolver (80.43\%). Unlike EncNet\cite{Zhang_2018_CVPR} which shows inconsistent performance across categories, ArmFormer maintains consistently high accuracy across all weapon types, critical for comprehensive threat assessment. The comparison with Segmenter\cite{strudel2021segmenter}, another transformer-based approach, validates our CBAM integration strategy, with ArmFormer achieving 49.18\% mIoU improvement due to domain-specific attention mechanisms.

The compact footprint of 4.886G FLOPs and 3.66M parameters positions ArmFormer as the optimal solution for edge deployment on resource-constrained devices including embedded GPUs, mobile processors, and AI accelerators in surveillance infrastructure. Unlike heavyweight models requiring server-grade hardware or ultra-lightweight models that sacrifice accuracy, ArmFormer achieves the critical balance necessary for practical real-world deployment in distributed security systems.

\subsection{Qualitative Results Analysis}

\begin{figure*}[t]
\centering
\includegraphics[width=\textwidth]{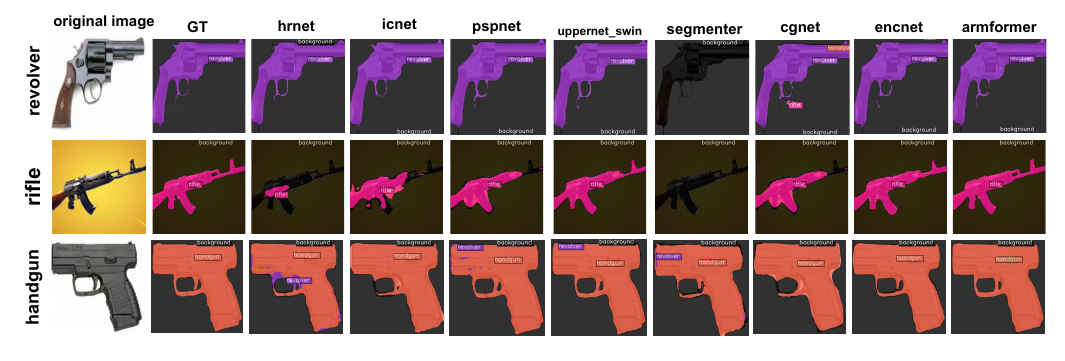}
\caption{Qualitative comparison of segmentation results across different models on challenging weapon detection scenarios. The figure presents three representative test cases: revolver detection (top row), rifle segmentation (middle row), and handgun identification (bottom row). Each column represents predictions from different models (HrNet\cite{SunXLW19}, ICNet\cite{zhao2018icnet}, PspNet\cite{zhao2017pspnet}, Uppernet\_swin\cite{xiao2018unified}, Segmenter\cite{strudel2021segmenter}, CGNet\cite{wu2020cgnet}, EncNet\cite{Zhang_2018_CVPR}, and our ArmFormer along with the original image). Ground truth annotations and background regions are shown with corresponding color coding for each weapon category. ArmFormer (rightmost column) consistently produces the most accurate pixel-level segmentation with precise object boundaries and minimal false predictions across all weapon categories.}
\label{fig:qualitative_results}
\end{figure*}

Figure \ref{fig:qualitative_results} presents qualitative segmentation results that visually demonstrate the superior performance of our proposed ArmFormer compared to state-of-the-art baselines across diverse weapon detection scenarios. The visual comparison reveals critical insights into model capabilities for handling real-world security challenges.

In the revolver detection case (Figure \ref{fig:qualitative_results} top row), ArmFormer produces the most accurate segmentation with complete object coverage and precise boundary delineation. While models like HrNet\cite{SunXLW19}, ICNet\cite{zhao2018icnet}, and PspNet\cite{zhao2017pspnet} successfully detect the revolver, they exhibit incomplete segmentation with missing regions and fragmented predictions. Notably, Segmenter\cite{strudel2021segmenter} completely fails to detect the revolver, producing an entirely black mask, which aligns with its extremely poor quantitative performance. Our ArmFormer achieves pixel-perfect segmentation with coherent object structure, demonstrating superior performance on the revolver class.

For rifle detection (Figure \ref{fig:qualitative_results} middle row), the qualitative results definitively establish ArmFormer's superiority. The rifle case is particularly challenging due to the elongated object structure and potential occlusions. While EncNet\cite{Zhang_2018_CVPR} and Uppernet\_swin\cite{xiao2018unified} show reasonable detection, they produce noisy predictions with boundary artifacts. Segmenter\cite{strudel2021segmenter} again demonstrates catastrophic failure with minimal detection capability. CGNet\cite{wu2020cgnet}, despite its high inference speed, shows incomplete segmentation with missing rifle components. ArmFormer produces the most accurate and complete rifle segmentation with clean boundaries and comprehensive object coverage, directly correlating with its superior quantitative performance compared to competitors.

In the handgun detection case (Figure \ref{fig:qualitative_results} bottom row), ArmFormer continues to demonstrate exceptional segmentation quality with precise boundary localization and complete object coverage. Most competing method is EncNet\cite{Zhang_2018_CVPR} which achieve reasonable handgun detection but suffer from boundary imprecision and occasional fragmentation. Uppernet\_swin\cite{xiao2018unified} shows competitive performance with relatively clean segmentation. However, Segmenter\cite{strudel2021segmenter} maintains its poor performance pattern, failing to provide meaningful segmentation results. CGNet\cite{wu2020cgnet} produces acceptable results but with some boundary artifacts. ArmFormer's handgun segmentation exhibits the highest fidelity with sharp, accurate boundaries and consistent object structure, reinforcing its robust performance across all weapon categories.

The qualitative analysis reveals a consistent pattern: ArmFormer maintains superior segmentation quality across all weapon categories and complexity levels. Unlike baseline models that show inconsistent performance across different scenarios, our proposed architecture demonstrates robust generalization enabled by strategic CBAM integration and transformer-based global context modeling. The visual results validate that ArmFormer's architectural design effectively addresses the fundamental challenges in weapon segmentation, including multi-scale feature representation, precise boundary localization, and reliable multi-class discrimination essential for security-critical applications.

\section{Ablation Study}

We conducted three comprehensive ablation studies on our proposed ArmFormer model to validate the effectiveness of our design choices and demonstrate the superiority of our approach. The ablation studies systematically evaluate: (1) the impact of FPN neck integration for multi-scale feature fusion, (2) the effect of lightweight CBAM configurations for computational efficiency, and (3) the influence of adaptive stage-specific CBAM parameters.

\begin{table}[htbp]
\centering
\caption{Ablation Study Results: ArmFormer vs. Alternative Configurations}
\label{tab:ablation_study}
\footnotesize
\begin{adjustbox}{width=\columnwidth,center}
\begin{tabular}{lccc}
\toprule
\textbf{Model Configuration} & \textbf{mIoU (\%)} & \textbf{mFscore (\%)} & \textbf{FPS} \\
\midrule
\rowcolor{lightgray}
\textbf{ArmFormer} & \textbf{80.64} & \textbf{89.13} & 82.26 \\
w/ FPN Neck & 79.92 & 88.70 & 78.85 \\
w/ Lightweight CBAM & 79.74 & 88.57 & \textbf{84.28} \\
w/ Adaptive CBAM & 78.05 & 87.54 & 81.15 \\
\bottomrule
\end{tabular}
\end{adjustbox}
\end{table}

\begin{table}[htbp]
\centering
\caption{Per-Class IoU Performance (\%)}
\label{tab:per_class_iou}
\footnotesize
\begin{adjustbox}{width=\columnwidth,center}
\begin{tabular}{lcccccc}
\toprule
\textbf{Configuration} & \textbf{Bg} & \textbf{Gun} & \textbf{Human} & \textbf{Knife} & \textbf{Rifle} & \textbf{Rev} \\
\midrule
\rowcolor{lightgray}
\textbf{ArmFormer} & 89.29 & \textbf{82.24} & 67.22 & \textbf{80.81} & \textbf{83.87} & \textbf{80.43} \\
w/ FPN Neck & 89.01 & 79.65 & \textbf{67.97} & 81.60 & 82.38 & 78.94 \\
w/ Lightweight & \textbf{89.46} & 79.88 & 66.90 & 82.87 & 81.32 & 78.04 \\
w/ Adaptive & 87.20 & 79.30 & 66.77 & 79.95 & 76.52 & 78.53 \\
\bottomrule
\end{tabular}
\end{adjustbox}
\end{table}

\begin{figure}[htbp]
\centering
\includegraphics[width=0.9\columnwidth]{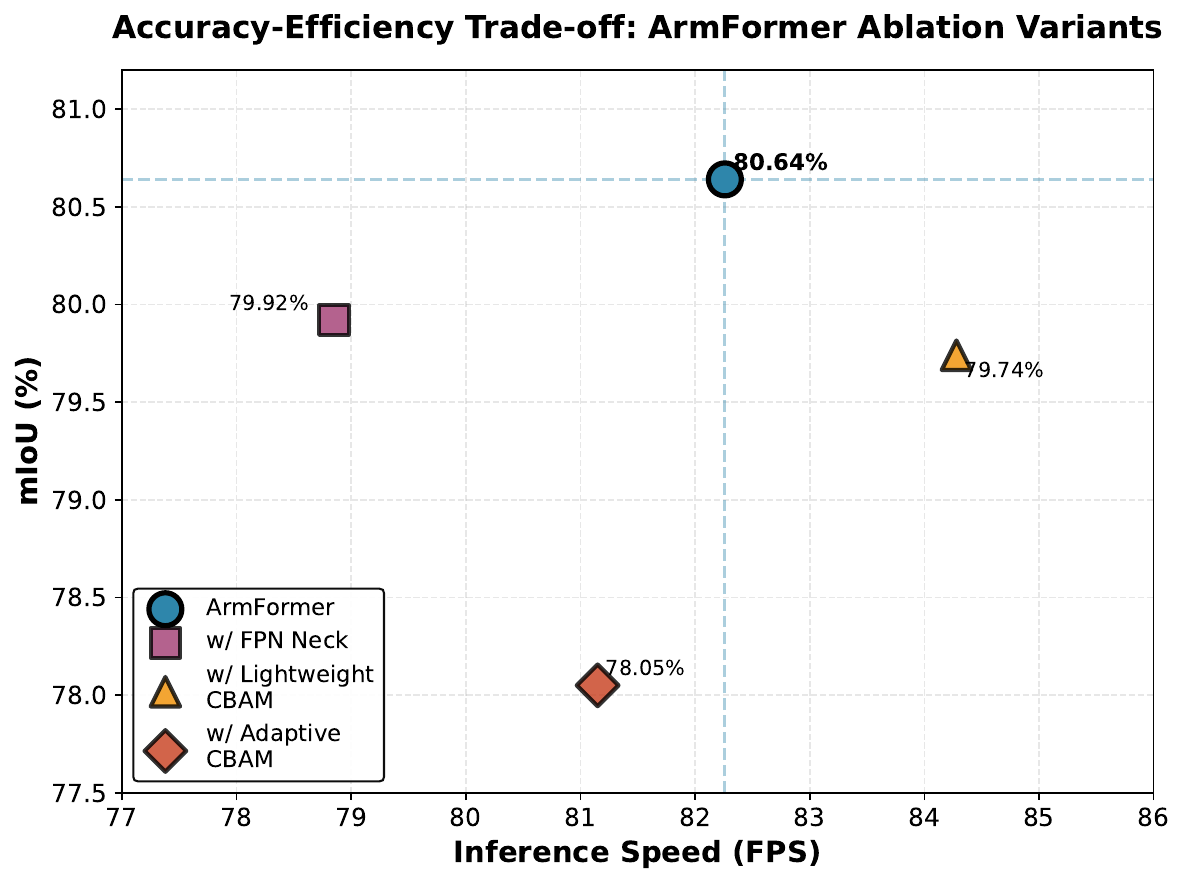}
\caption{Accuracy-efficiency trade-off visualization for ArmFormer ablation variants. The scatter plot illustrates the relationship between inference speed (FPS) and segmentation accuracy (mIoU) across different configurations. Our base ArmFormer (blue circle) achieves the optimal balance, positioned in the upper-right region with highest mIoU (80.64\%) and competitive FPS (82.26). The dashed reference lines intersect at ArmFormer's performance point, demonstrating its superior positioning in the accuracy-efficiency space.}
\label{fig:ablation_graph}
\end{figure}

Our comprehensive ablation studies demonstrate the superiority of our proposed ArmFormer design and validate our architectural choices. The results clearly show that our base ArmFormer achieves the best overall performance with 80.64\% mIoU and 89.13\% mFscore, outperforming all ablation variants. Figure \ref{fig:ablation_graph} visualizes the accuracy-efficiency trade-off across all configurations, revealing critical insights into the design space exploration.

\textbf{FPN Neck Integration Analysis.} As illustrated in Figure \ref{fig:ablation_graph}, the FPN neck variant (red square) falls into the lower-left quadrant, exhibiting both reduced accuracy and slower inference. Adding FPN neck integration results in a performance decline from 80.64\% to 79.92\% mIoU (-0.72\%) and reduced inference speed (82.26 to 78.85 FPS). While FPN theoretically provides better multi-scale feature fusion, our results demonstrate that the additional computational overhead outweighs the benefits, validating our decision to exclude FPN from the main ArmFormer architecture. The graph clearly shows that this configuration moves away from the optimal trade-off point.

\textbf{Lightweight CBAM Configuration.} The lightweight variant (orange triangle) appears in the upper-right region of Figure \ref{fig:ablation_graph}, achieving the highest FPS (84.28) among all configurations. This variant with higher reduction ratios (32) and smaller kernels (3) achieves 79.74\% mIoU while improving computational efficiency. Although this configuration offers better speed performance, it sacrifices 0.90\% mIoU compared to our main model. The visualization reveals that while lightweight CBAM provides marginal speed gains (+2.02 FPS), the accuracy degradation makes it suboptimal for security-critical applications where precision is paramount. Our balanced CBAM parameters (reduction ratio: 16, kernel size: 7) provide the optimal accuracy-efficiency trade-off, as evidenced by ArmFormer's superior positioning in the accuracy-efficiency space.

\textbf{Adaptive CBAM Parameters.} The adaptive configuration (brown diamond) demonstrates the worst performance in Figure \ref{fig:ablation_graph}, positioned in the lower region with the lowest mIoU (78.05\%, -2.59\% degradation). This counterintuitive result indicates that overly complex adaptive mechanisms can hinder optimization and lead to suboptimal performance, particularly evident in rifle detection (83.87\% → 76.52\%). The graph visualization definitively shows that adaptive stage-specific parameters do not improve the accuracy-efficiency trade-off, validating our design choice of uniform CBAM parameters across stages. The performance point's position away from the optimal frontier confirms that architectural simplicity combined with strategic attention placement yields superior results.

Our ablation studies conclusively prove that the proposed ArmFormer with uniform CBAM integration achieves superior performance compared to all alternative configurations. The consistent attention mechanisms across stages, optimal CBAM parameters, and streamlined architecture without FPN neck represent the most effective design for weapon detection segmentation.

\section{Conclusion}

This paper presented ArmFormer, a lightweight transformer-based semantic segmentation framework for real-time multi-class weapon detection in edge environments. By strategically integrating CBAM with MixVisionTransformer architecture, ArmFormer achieves state-of-the-art performance with 80.64\% mIoU and 89.13\% mFscore at 82.26 FPS, outperforming heavyweight models requiring up to 48× more computation. With only 4.886G FLOPs and 3.66M parameters, ArmFormer is optimally suited for deployment on portable security cameras, programmable surveillance systems, autonomous drones, and embedded AI accelerators, enabling real-time threat assessment without cloud dependency. Comprehensive ablation studies validate that uniform CBAM integration across encoder stages yields superior accuracy-efficiency trade-offs. Despite these achievements, limitations include performance variability under extreme lighting conditions, challenges with heavily occluded weapons, and the need for larger-scale diverse training datasets. Future work will explore model quantization and pruning for ultra-low-power scenarios, multi-modal fusion with thermal and depth sensors for enhanced detection under challenging conditions, real-time video stream processing with temporal consistency, and federated learning strategies for privacy-preserving collaborative training across distributed edge devices.

% \section*{Acknowledgment}

% The authors would like to thank the anonymous reviewers for their valuable feedback and suggestions that helped improve this work.

\bibliographystyle{IEEEtran}
\bibliography{armformer_ref}

\end{document}